# Occluded object reconstruction for first responders with augmented reality glasses using conditional generative adversarial networks

Kyongsik Yun*, Thomas Lu, Edward Chow

Jet Propulsion Laboratory, California Institute of Technology

4800 Oak Grove Drive, Pasadena, CA 91109

## ABSTRACT

Firefighters suffer a variety of life-threatening risks, including line-of-duty deaths, injuries, and exposures to hazardous substances. Support for reducing these risks is important. We built a partially occluded object reconstruction method on augmented reality glasses for first responders. We used a deep learning based on conditional generative adversarial networks to train associations between the various images of flammable and hazardous objects and their partially occluded counterparts. Our system then reconstructed an image of a new flammable object. Finally, the reconstructed image was superimposed on the input image to provide "transparency". The system imitates human learning about the laws of physics through experience by learning the shape of flammable objects and the flame characteristics.

**Keywords:** partially occluded object reconstruction, augmented reality, generative adversarial networks, machine learning, computer vision

## 1. INTRODUCTION

More than 30,000 firefighters are injured each year during firefighting operations [1,2]. Slips, trips, and falls cause a large number of firefighter injuries, reinforcing the need for advanced ground support for firefighters [3,4]. As part of the Assistant for Understanding Data through Reasoning, Extraction, and sYnthesis (AUDREY) project, supported by the Department of Homeland Security (DHS) [5-7], we aim to provide situational awareness information to next-generation first responders to help them perform safe, healthy, and successful missions.

Wearable sensors and augmented reality glasses were developed to provide more data to first responders [7,8]. However, more data is not necessarily helping first responders do their jobs. Overwhelming data may actually prevent the first responders from performing important activities [9,10]. A first responder cannot make prompt decisions because he/she cannot quickly extract key insights because of too much information. This is not because there is no relevant data. Information overload is a barrier that prevents the first responder from performing his or her work safely and efficiently.

It is important to manage the cognitive load of the first responder. Human working memory holds information for only a few seconds and holds only five to seven items at a time [11,12]. Working memory can even manipulate information [13,14]. Therefore, there is a desperate need for an automated system that can pick up important data and translate it into practical knowledge or insight. In the first responder scenario, it is important to process the information in real time and selectively present it to the first responder. The information must be relevant to the situation and provide insight.

One of the most important pieces of information in augmented reality glasses is the "see-through" feature [15,16]. If firefighters can see through the fire, it will help them to navigate the incident more safely and efficiently. Moreover, firefighters can quickly find victims partially occluded by objects. Estimating the size and shape of a fire in a partially occluded situation is particularly important in that it can predict a sudden fire explosion (flashover phenomenon) [17]. If firefighters are provided with see-through capability, they can be safe from dangerous environments and fully understand the surrounding situation.

See-through functionality has been studied in a variety of areas, including user interfaces [18], mixed reality display devices [19], and robotic networks [20]. Lai and colleagues transformed an object into another object using deformable optics that can be applied to see-through walls and cloaking [21]. Pseudo-transparency was suggested using an additional imaging device at the occluded location [22]. Previous studies on see-through functionality required additional cameras or special types of optics. Semantic occluded object reconstruction inspired by human vision can be a solution without additional hardware.

*kyun@jpl.nasa.gov; phone 1 818 354-1468; fax 1 818 393-6752; jpl.nasa.gov

In this study, we built a partially occluded object reconstruction method for augmented reality glasses for first responders.

## 2. CONDITIONAL GENERATIVE ADVERSARIAL NETWORKS

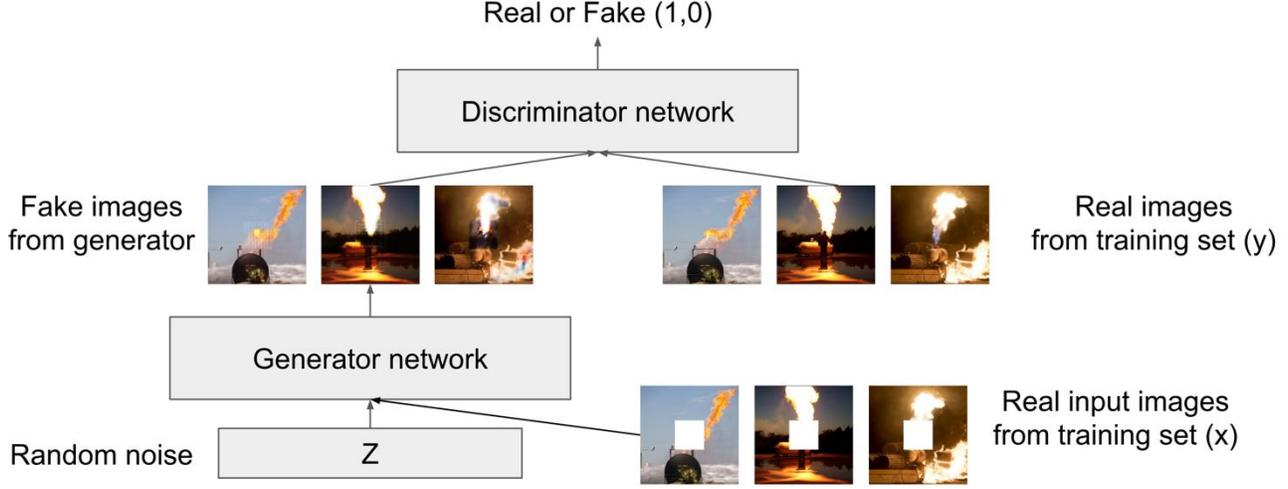

Figure 1. Conditional generative adversarial networks architecture. The generator network uses the occluded real image as input to output a reconstructed fake image. The generated image is compared with the real image in the discriminator network that outputs the classification of the real or the fake. With adversarial training, the generator produces a more realistic image, and the discriminator more accurately distinguishes between real and fake.

We used generative adversarial networks to train associations between various images of flammable and hazardous objects and their partially occluded counterparts. Generative adversarial networks learn the loss of classifying whether the output image is real, while at the same time the networks learn the generative model to minimize this loss [23]. To train the generative adversarial networks, the objective is to solve the min-max game, finding the minimum over $\theta_g$, or the parameters of our generator network G and maximum over $\theta_d$, or the parameters of our discriminator network D.

$$\min_{\theta_g} \max_{\theta_d} \left[ \mathbb{E}_{y \sim p_{data}} \log D_{\theta_d}(y) + \mathbb{E}_{z \sim p(z)} \log \left( 1 - D_{\theta_d}\left(G_{\theta_g}(z)\right) \right) \right] \quad (1)$$

The first term in the objective function (1) is the likelihood, or expectation, of the real data being real from the data distribution $P_{data}$. The log $D(y)$ is the discriminator output for real data *y*. If *y* is real, *D(y)* is 1 and log *D(y)* is 0, which becomes the maximum. The second term is the expectation of *z* drawn from *P(z)*, meaning the random data input for our generator network (Figure 1). *D(G(z))* is the output of our discriminator for generated fake data *G(z)*. If *G(z)* is close to real, *D(G(z))* is close to 1, and the *log (1-D(G(z)))* becomes very small (minimized).

$$\min_{\theta_g} \max_{\theta_d} \left[ \mathbb{E}_{y \sim p_{data}} \log D_{\theta_d}(y) + \mathbb{E}_{z \sim p(z), x \sim p(x)} \log \left( 1 - D_{\theta_d}\left(G_{\theta_g}(z, x)\right) \right) \right] \quad (2)$$

$$\min_{\theta_g} \max_{\theta_d} \left[ \mathbb{E}_{y \sim p_{data}, x \sim p_{data}} \log D_{\theta_d}(y, x) + \mathbb{E}_{z \sim p(z), x \sim p(x)} \log \left(1 - D_{\theta_d}\left(G_{\theta_g}(z, x), x\right)\right) \right] \quad (3)$$

We used the random input variable *z* as input to the generator network. An image can be used as input to the generator network instead of *z* (Figure 1). In the objective function (2), *x*, the real input image was a conditional term for our generator. We then added a conditional term *x* to our discriminator network, as in function (3). The conditional adversarial networks learned the mapping from the input image to the output image and learn the loss function [24,25].

Figure 1 showed the overall architecture, including input occluded images, generated output reconstruction images, and ground truth images. While the original generative adversarial networks generate real-looking images from the random noise input *z*, the conditional generative adversarial networks can find the association of the two images (i.e., original and occluded), and transform or reconstruct the real input image into another image.

Our goal is to reconstruct partially occluded objects in the image, so it can be helpful to train the basic structure of the actual objects (fire, gas tank, roof, etc.) and to maintain the structural similarity of the input and output images. Therefore, we used U-Net architecture [26] for our generator based on an encoder-decoder network that was gradually downsampled and upsampled for efficiency. The network then applied the skip layer. That is, each downsampling layer is sent to and connected to the corresponding upsampling layer. Finally, the upsampling layer could directly learn important structural features from the downsampling layer.

## 3. TRAINING AND TESTING

We collected 100 images from Google Images (keywords: gas tank fire) (Figure 2). We then added artificial occlusion to the image and used it as input. We used 80 image pairs randomly selected for training and 20 image pairs for testing.

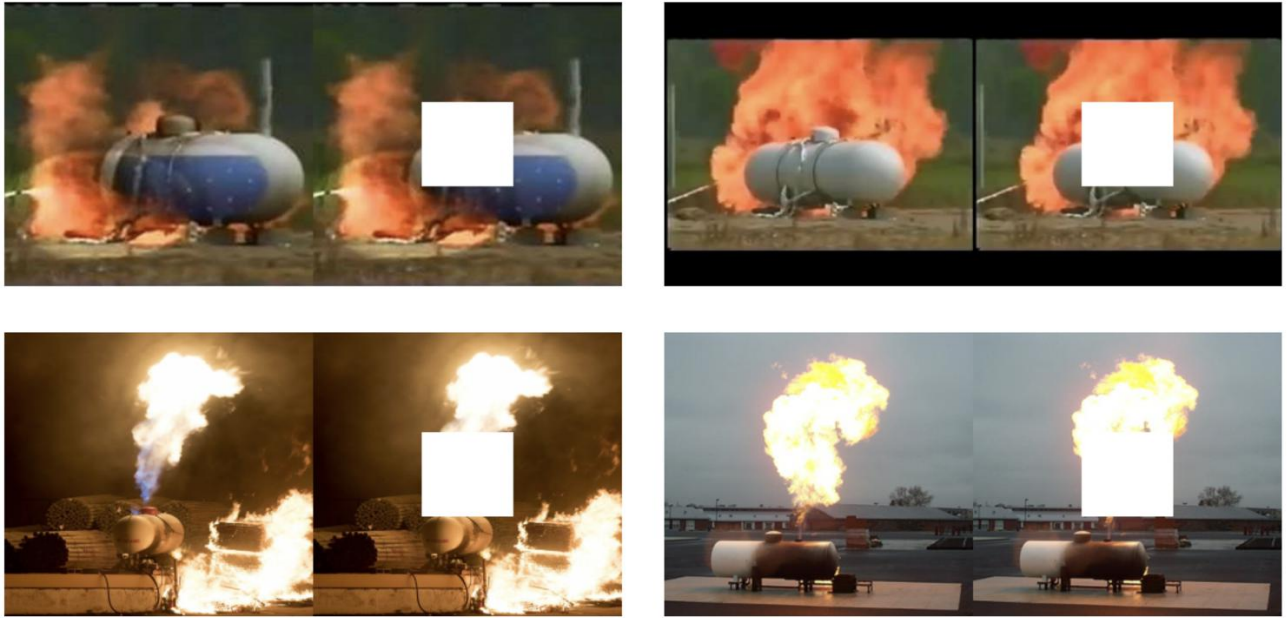

Figure 2. Representative training image pairs for our conditional generative adversarial networks. Propane gas tank images in various situations were used for training. The original image was entered into the discriminator network and the occluded copy was used as the generator network input.

After training conditional generative adversarial networks, our system reconstructed test images of partially occluded flammable objects. Finally, the reconstructed image was superimposed on the input image to provide "transparency" (Figure 3). Representative reconstructions are shown in Figure 4. Five out of 20 test images were incorrectly reconstructed as shown in Figure 4C (25% error rate).

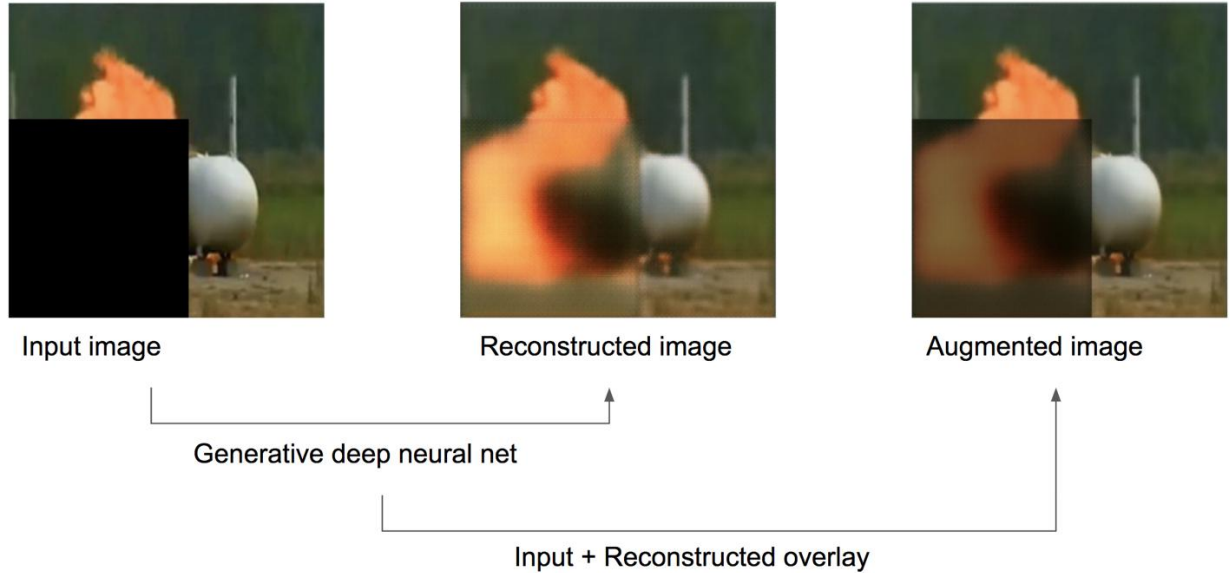

Figure 3. Reconstructing a partially occluded object for first responders using conditional generative adversarial networks and augmented reality glasses. The reconstructed image was overlaid on the input image and used as a visual enhancement feature for the first responder.

## 4. DISCUSSION

Conditional generative adversarial networks have shown good results on small data sets (fewer than 100 samples) [17,24]. The advantage of a small datasets is the speed of training. The findings of the current study were based on less than five hours of training on a single NVIDIA GTX1080 GPU. Because the time it takes to convert the test output image is shorter than 200 milliseconds, this study can be applied as a real solution that incorporates improved insight into the augmented reality glasses of the next generation first responder.

Through experience, humans learn the general form of an object, then reconstruct and visualize a partially occluded object internally. Each brain area in the visual cortex processes different contextual levels of information. Lower-level visual cortical regions (V1, V2) detect oriented edges (lines and curves), and then V4 integrates information to identify components of an object (eyes, nose, etc.). Finally, posterior and anterior inferior temporal cortices process semantic information [27]. This is called the brain's bottom-up object recognition [28]. On the other hand, the human brain can reverse the bottom-up process to visualize a new object based on semantic information, or to visualize the occluded part of an object, which is called the top-down process. Conditional generative adversarial networks can be thought of as an implementation of a human brain process that combines a bottom-up recognition (discriminator network) and a top-down imagination (generator network).

Previous studies have attempted to reconstruct an occluded object in augmented reality by solving mathematical models [16,29]. Using deep learning-based conditional generative adversarial networks, we could consider complex backgrounds and multiple objects more easily than mathematical models. Deep learning can be thought of as a nonlinear numerical model. Mathematical models may not be accurate in a complex real world.

If the objective is to provide a warning and help identify the victim in a fire situation, one may argue that we can provide the necessary specific information directly, without having to reconstruct the partially occluded object as in this study. It may be true, but by providing a reconstruction of the occluded object, firefighters can discover other significant risks that cannot be identified by a warning algorithm. Moreover, in terms of explanatory and transparent solutions, lower level information (visualization of occluded parts) is more intuitive and faster than warning signals [30-32].

The reason for not reconstructing the objects correctly in some test images (Figure 4C) may be due to a lack of variability in the training data set. In the next study, we will increase the number and variability of training images to deal with various types of gas tanks and fire events. Contextual information, including building structure, fire cause, and gas / chemical sensor information, will help provide more accurate semantic reconstruction. Image changes over time and perspective changes can be added to increase accuracy.

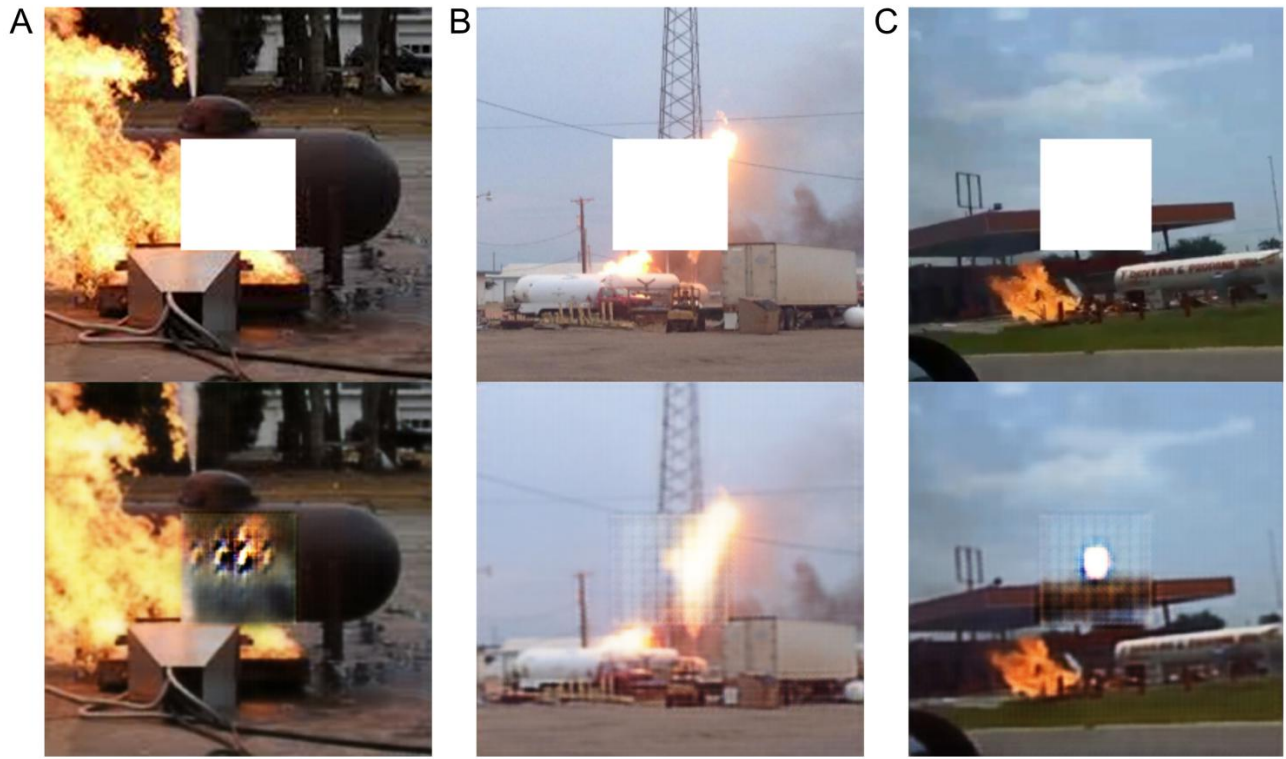

Figure 4. Representative correct and incorrect reconstruction examples. (A) The fire pattern was not generated correctly. (B) The fire pattern was accurately generated. The transmission tower located in the background behind the fire was not reconstructed. (C) A false fire pattern was created (negative example). The roof was generated correctly.

## 5. CONCLUSION

This system imitates human learning about the laws of physics through experience by learning the shape and flame characteristics of flammable objects. This system will help the first responders to strengthen their vision in a fire-fighting situation. Augmented reality glasses with our occluded object reconstruction technique will provide an important support to the next generation first responders.

## ACKNOWLEDGMENT

The research was carried out at the Jet Propulsion Laboratory, California Institute of Technology, under a contract with the National Aeronautics and Space Administration. The research was funded by the U.S. Department of Homeland

# REFERENCES

[1] Bowyer, M. E., Miles, V., Baldwin, T. N. & Hales, T. R., "Preventing deaths and injuries of fire fighters during training exercises", (2016).
[2] Butler, C., Marsh, S., Domitrovich, J. W. & Helmkamp, J., "Wildland firefighter deaths in the United States: A comparison of existing surveillance systems", Journal of occupational and environmental hygiene, 14, 258-270 (2017).
[3] Karter, M. J. & Molis, J. L., US firefighter injuries-2012. (National Fire Protection Association, Fire Analysis and Research Division, 2013).
[4] Smith, D. L., Firefighter fatalities and injuries: the role of heat stress and PPE. (Firefighter Life Safety Research Center, Illinois Fire Service Institute, University of Illinois, 2008).
[5] Mattson, P., Merrill, J., Lustig, T. & Deso, W. in *Performance of Protective Clothing and Equipment: 10 th Volume, Risk Reduction Through Research and Testing* (ASTM International, 2016).
[6] Baggett, R. K. & Foster, C. S., "TheFutureofHomelandSecurity Technology", Homeland Security Technologies for the 21st Century, 257 (2017).
[7] Sanquist, T. & Brisbois, B., "Attention and Situational Awareness in First Responder Operations", (2016).
[8] Wu, X., Dunne, R., Zhang, Q. & Shi, W., "Edge Computing Enabled Smart Firefighting: Opportunities and Challenges", (2017).
[9] Abu-Elkheir, M., Hassanein, H. S. & Oteafy, S. M., in Wireless Communications and Mobile Computing Conference (IWCMC), 2016 International, 188-193 (IEEE).
[10] Cinque, M. *et al.*, "SECTOR: Secure Common Information Space for the Interoperability of First Responders", Procedia Computer Science, 64, 750-757 (2015).
[11] Anderson, J. R., Reder, L. M. & Lebiere, C., "Working memory: Activation limitations on retrieval", Cognit. Psychol., 30, 221-256 (1996).
[12] Callicott, J. H. *et al.*, "Physiological characteristics of capacity constraints in working memory as revealed by functional MRI", Cereb. Cortex, 9, 20-26 (1999).
[13] Crone, E. A., Wendelken, C., Donohue, S., van Leijenhorst, L. & Bunge, S. A., "Neurocognitive development of the ability to manipulate information in working memory", Proceedings of the National Academy of Sciences, 103, 9315-9320 (2006).
[14] Rypma, B. & D'Esposito, M., "The roles of prefrontal brain regions in components of working memory: effects of memory load and individual differences", Proceedings of the National Academy of Sciences, 96, 6558-6563 (1999).
[15] Marescaux, J., Rubino, F., Arenas, M., Mutter, D. & Soler, L., "Augmented-reality–assisted laparoscopic adrenalectomy", JAMA, 292, 2211-2215 (2004).
[16] Azuma, R. *et al.*, "Recent advances in augmented reality", IEEE computer graphics and applications, 21, 34-47 (2001).
[17] Yun, K., Bustos, J. & Lu, T., "Predicting Rapid Fire Growth (Flashover) Using Conditional Generative Adversarial Networks", arXiv:1801.09804, (2018).
[18] Bier, E. A., Stone, M. C., Pier, K., Buxton, W. & DeRose, T. D., in Proceedings of the 20th annual conference on Computer graphics and interactive techniques, 73-80 (ACM).
[19] Perez, K. S., Vaught, B. I., Lewis, J. R., Crocco, R. L. & Kipman, A. A.-A. (Google Patents, 2016).
[20] Mostofi, Y., "Cooperative wireless-based obstacle/object mapping and see-through capabilities in robotic networks", IEEE Transactions on Mobile Computing, 12, 817-829 (2013).
[21] Lai, Y. *et al.*, "Illusion optics: the optical transformation of an object into another object", Physical review letters, 102, 253902 (2009).
[22] Wigdor, D., Forlines, C., Baudisch, P., Barnwell, J. & Shen, C., in Proceedings of the 20th annual ACM symposium on User interface software and technology, 269-278 (ACM).
[23] Goodfellow, I. *et al.*, in Advances in neural information processing systems, 2672-2680.

[24] Isola, P., Zhu, J.-Y., Zhou, T. & Efros, A. A., "Image-to-image translation with conditional adversarial networks", arXiv preprint arXiv:1611.07004, (2016).
[25] Mirza, M. & Osindero, S., "Conditional generative adversarial nets", arXiv preprint arXiv:1411.1784, (2014).
[26] Ronneberger, O., Fischer, P. & Brox, T., in International Conference on Medical Image Computing and Computer-Assisted Intervention, 234-241 (Springer).
[27] Hubel, D. H. & Wiesel, T., "Shape and arrangement of columns in cat's striate cortex", The Journal of physiology, 165, 559-568 (1963).
[28] Yun, K. & Stoica, A., in Systems, Man, and Cybernetics (SMC), 2016 IEEE International Conference on, 002220-002223 (IEEE).
[29] Lee, J., Hirota, G. & State, A., "Modeling real objects using video see-through augmented reality", Presence: Teleoperators & Virtual Environments, 11, 144-157 (2002).
[30] Gunning, D., "Explainable artificial intelligence (xai)", Defense Advanced Research Projects Agency (DARPA), nd Web, (2017).
[31] Pei, K., Cao, Y., Yang, J. & Jana, S., in Proceedings of the 26th Symposium on Operating Systems Principles, 1-18 (ACM).
[32] Samek, W., Wiegand, T. & Müller, K.-R., "Explainable Artificial Intelligence: Understanding, Visualizing and Interpreting Deep Learning Models", arXiv preprint arXiv:1708.08296, (2017).